\documentclass[robotics,article,submit,moreauthors,pdfLaTeX]{Definitions/mdpi} 
\firstpage{1} 
\pubvolume{1}
\issuenum{1}
\articlenumber{0}
\pubyear{2024}
\copyrightyear{2024}
\datereceived{ } 
\daterevised{ } 
\dateaccepted{ } 
\datepublished{ } 
\hreflink{https://doi.org/} 
\usepackage[utf8]{inputenc}
\usepackage{amsmath,amssymb,amsfonts}
\usepackage{bm}
\usepackage{url}
\usepackage{graphicx,graphics}
\usepackage{multirow}
\usepackage{booktabs}
\usepackage{textcomp}
\usepackage{tabularx}
\usepackage{relsize}
\usepackage[flushleft]{threeparttable}
\usepackage{rotating}
\usepackage{comment}
\usepackage{soul}
\usepackage{tikz}
\usepackage{algpseudocode}  
\usepackage{subcaption}

\newcommand{\midsepremove}{\aboverulesep = 0.2mm \belowrulesep = 0.2mm}
\newcommand{\midsepdefault}{\aboverulesep = 0.605mm \belowrulesep = 0.984mm}
\Title{Multimodal Control of Manipulators:
Coupling Kinematics and Vision for
Self-Driving Laboratory Operations 
}

\TitleCitation{Title}

\Author{Shifa Sulaiman $^{1,*}$\orcidA{}, Amarnath A H $^{2}$, Simon Bøgh $^{1}$\orcidB{}, and Naresh Marturi $^{3}$\orcidC{}}

\AuthorNames{Shifa Sulaiman, Amarnath A H and Naresh Marturi}

\AuthorCitation{Sulaiman, S.; A H, A.; Marturi, N.}

\address{%
$^{1}$ \quad Department of Electronics Systems, Aalborg University, Denmark; 
$^{2}$ \quad Nawe Robotics, Calicut, India; 
$^{3}$ \quad Extreme Robotics Laboratory, School of Metallurgy \& Materials, University of Birmingham,
Birmingham, United Kingdom; 
}

\corres{Correspondence: ssajmech@gmail.com} 

\abstract{
 Self-driving laboratories are redefining autonomous experimentation by integrating robotic manipulation, computer vision, and intelligent planning to accelerate scientific discovery. This work presents a vision-guided motion planning framework for robotic manipulators operating in dynamic laboratory environments, with a focus on evaluating motion smoothness and control stability. The framework enables autonomous detection, tracking, and interaction with textured objects through a hybrid scheme that couples advanced motion planning algorithms with real-time visual feedback.
 Kinematic modeling of the manipulator is carried out using the screw theory formulations, which provides a rigorous foundation for deriving forward kinematics and the space Jacobian. These formulations are further employed to compute inverse kinematic solutions via the Damped Least Squares (DLS) method, ensuring stable and continuous joint trajectories even in the presence of redundancy and singularities. Motion trajectories toward target objects are generated using the RRT* algorithm, offering optimal path planning under dynamic constraints.
Object pose estimation is achieved through a vision pipeline that integrates feature-based detection with homography-driven depth analysis, enabling adaptive tracking and dynamic grasping of textured objects. The manipulator’s performance is quantitatively evaluated using smoothness metrics, RMSE pose errors, and joint motion profiles including velocity continuity, acceleration, jerk, and snap.
Simulation studies demonstrate the robustness and adaptability of the proposed framework in autonomous experimentation workflows, highlighting its potential to enhance precision, scalability, and efficiency in next-generation self-driving laboratories.}

\keyword{self-driving laboratory; vision-based object tracking, Jacobian inverse kinematics, Trajectory planning (RRT*), Autonomous robotic manipulation} 
\begin{document}
\section{Introduction}
\label{sec:intro}
Self-driving laboratories (SDL) are transforming the landscape of autonomous experimentation by integrating robotics, computer vision, and intelligent planning to accelerate scientific workflows. Within these environments, dexterous manipulators equipped with sophisticated grippers play a pivotal role in executing precise and adaptive tasks, often outperforming human operators in terms of reliability and efficiency \cite{alexandra2023new} - \cite{sultanov2023virtual}. A key challenge in such dynamic settings lies in enabling robotic systems to interact with textured objects in real time, requiring seamless coordination between vision-based perception and motion planning. This coordination must not only ensure accurate object tracking but also generate smooth and feasible joint trajectories for safe and efficient manipulation.

This work presents a foundational control framework designed to address this challenge by coupling real-time object tracking with smooth and stable motion execution. The system integrates a hybrid vision pipeline based on feature detection and homography-driven pose estimation with Jacobian-based motion planning for a 7-DOF manipulator. While the vision module is validated using a highly-textured, opaque object to ensure reliable tracking, the primary contribution lies in the rigorous quantitative analysis of joint motion dynamics. Specifically, we evaluate velocity continuity, acceleration, jerk, snap, and smoothness cost functions to benchmark trajectory quality and mechanical stability.
Rather than claiming architectural novelty, this study focuses on validating the control pipeline’s ability to generate smooth, feasible trajectories in response to dynamic visual input. Depth information is leveraged to interpret object orientation in 3D space, guiding the manipulator’s end-effector toward dynamically changing targets. Trajectories are generated using the RRT* algorithm, while inverse kinematic solutions are computed using Damped Least Squares (DLS), chosen for its potential to optimize joint smoothness and precision \cite{sulaiman2025jacobian}. By establishing a modular and reproducible baseline, this framework lays the groundwork for future extensions involving more complex vision modalities and real-world SDL objects such as transparent glassware, reflective surfaces, and low-texture microplates.


The remainder of this paper is organized as follows. Section \ref{sec:bckgrnd} focuses on previous work related to kinematic analysis, motion planning schemes, and vision algorithms. Kinematic modeling and workspace analyses of the manipulator, motion planning schemes and vision algorithm are presented in Section \ref{sec:Methodologies}. Experimental results, including simulation and comparison studies of the proposed motion schemes, are presented in Section \ref{sec:experiments}.
\section{Related Work}
\label{sec:bckgrnd}

In the context of SDLs, mobile manipulators that combine a dexterous arm with a mobile base are proving essential for automating complex experimental workflows. These integrated platforms offer both spatial mobility and fine-grained manipulation capabilities, allowing them to navigate dynamic lab environments and interact with diverse instruments and materials. Their deployment in chemical research settings introduces distinct challenges and opportunities, particularly due to the delicate and potentially hazardous nature of lab operations. Tasks such as handling fragile glassware, precisely dispensing reagents, and interfacing with analytical equipment require high levels of accuracy, repeatability, and safety making mobile manipulators a cornerstone of autonomous ssecientific discovery. A motion planning scheme associated with a manipulator typically involves several phases, including kinematic modeling, workspace analysis, trajectory planning, and collision avoidance methods. For a Jacobian-based motion planning scheme, a manipulator's kinematic model is essential to determine joint solutions. Kinematic equations for a manipulator can be derived using various techniques, such as the geometric approach, the DH method, the screw theory approach, and iterative methods. Kinematic modeling of a manipulator establishes the relationship between the pose of an end effector and its corresponding joint configurations. It also defines the mapping between the linear and angular velocities of the end effector and the manipulator's joints.

Screw theory formulations are widely used for the kinematic modeling of robotic systems with higher number of degrees-of-freedom (dof). This approach proved to be particularly flexible for modeling complex systems with coupled and offset joints. Liao et al. \cite{liao2020novel} demonstrated the application of screw theory formulations in determining inverse kinematic solutions for a 6-DOF manipulator with offset joints . They utilized the Paden-Kahen (PK) method in conjunction with screw theory to establish the relationship between joint angular variations and end effector positions. Another approach to kinematic modeling, combining screw theory and PK methods, was presented by Gandhi et al. \cite{gandhi2021modeling}. They applied this methodology to model a 3-DOF street cleaning robotic manipulator, obtaining inverse solutions using the PK method. A comparison of joint solutions obtained through the PK method and numerical inverse kinematic approaches was conducted to assess computational efficiency. Additionally, Wang et al. \cite{wang2019kinematical} illustrated a derivation and solution strategy for solving kinematic equations of a free-floating robot arm. Their method involved a combined approach of screw theory and conservation of linear momentum theory. They employed Newton's iterative method during inverse kinematic analysis and compared results with other iterative methods to validate the accuracy of the approach.

Screw theory formulations implemented for determining inverse kinematic solutions for a 6 dof manipulator with offset joints was demonstrated by Liao et al. \cite{liao2020novel}. Paden-Kahen (PK) method was used along with screw theory approach to map a relation between joint angular variations and end effector positions. Another kinematic modeling approach combining screw theory and PK methods were given in \cite{gandhi2021modeling}. A 3 dof street cleaning robotic manipulator were modeled using screw theory approach and inverse solutions were obtained using PK method. Joint solutions determined using PK method were compared with joint solutions obtained using a numerical inverse kinematic approach to study a computational efficiency of the proposed approach. Wang et al. \cite{wang2019kinematical} illustrated a derivation and a solution strategy for solving kinematic equations of a free floating robot arm using a combined approach of screw theory method and conservation of linear momentum theory. Newton’s iterative method was also used during inverse kinematic analysis and results were compared with iterative methods to prove an accuracy of the method. 

Liu et al. \cite{liu2018trajectory} introduced a kinematic modeling approach for a 6-DOF industrial robot utilizing screw theory formulations. They employed a Particle Swarm Optimization (PSO)-based algorithm to minimize synthesis errors while considering kinematic and dynamic constraints. Another method for kinematic modeling of a redundant manipulator, combining screw theory with the Newton-Raphson method, was presented by Ge et al. \cite{ge2022kinematics}. They derived forward kinematic equations based on screw theory formulations and obtained joint solutions using the Newton-Raphson method. Screw theory-based kinematic modeling and motion analysis of a fixed-base dual-arm robot were demonstrated by Sulaiman et al. \cite{sulaiman2021modelling}. They utilized screw theory-derived kinematic equations to plot the robot's workspace. Additionally, Sulaiman et al. \cite{sulaiman2022modeling} derived kinematic equations for a 10-DOF dual-arm robot with a wheelbase using screw theory. These equations were employed to evaluate singularities and dexterous regions within the robot's workspace. An iterative method for determining forward kinematic equations using screw theory formulations was demonstrated by Medrano et al. \cite{medrano2022forward}. They applied this approach to model a 6-DOF manipulator and conducted simulation studies to demonstrate the advantages of their method. Sun et al. \cite{sun2021generalized} illustrated a kinematic modeling method based on screw theory applicable for hybrid mechanisms. They computed Jacobian and Hessian matrices using a recursive method and used the proposed approach to derive the kinematic model of a hybrid robot. Simulation results indicated the accuracy of the obtained kinematic equations, demonstrating the practical feasibility of the proposed method for kinematic analysis of hybrid mechanisms.

Recent advancements in vision-based grasping have significantly enhanced the capabilities of robotic manipulators in dynamic and unstructured environments. Hélénon et al. \cite{helenon2023toward} introduced a plug-and-play vision-based grasping framework that leverages Quality-Diversity (QD) algorithms to generate diverse sets of open-loop grasping trajectories. Their system integrates multiple vision modules for 6-DoF object detection and tracking, allowing trajectory generalization across different manipulators such as the Franka Research 3 and UR5 arms. This modular approach improves adaptability and reproducibility in robotic grasping tasks. Kushwaha et al. \cite{kushwaha2025vision} proposed a vision-based intelligent grasping system using sparse neural networks to reduce computational overhead while maintaining high grasp accuracy. Their Sparse-GRConvNet and Sparse-GINNet architectures utilize the Edge-PopUp algorithm to identify high-quality grasp poses in real time. Extensive experiments on benchmark datasets and the a cobot validated the models’ effectiveness in manipulating unfamiliar objects with minimal network parameters. Wang et al. \cite{wang2022polynomial} developed a trajectory planning method for manipulator grasping under visual occlusion using monocular vision and multi-layer neural networks. Their approach combines Gaussian sampling with Hopfield neural networks to optimize grasp paths in cluttered environments. The proposed method achieved a 99.5\% identification accuracy and demonstrated significant improvements in motion smoothness and efficiency.

Zhang et al. \cite{zhang2022robotic} presented a comprehensive survey of robotic grasping techniques, tracing developments from classical analytical methods to modern deep learning-based approaches. Their work highlights the evolution of grasp synthesis and the integration of vision algorithms in robotic manipulation. Similarly, Newbury et al. \cite{newbury2023deep} reviewed deep learning approaches to grasp synthesis, emphasizing the role of convolutional neural networks and transformer models in improving grasp reliability. Du et al. \cite{du2021vision} provided a detailed review of vision-based robotic grasping, covering object localization, pose estimation, and grasp inference for parallel grippers. Their analysis underscores the importance of combining RGB-D data with machine learning models to enhance grasp precision in real-world scenarios. These findings align with the growing trend of integrating vision and motion planning for dexterous manipulation. In addition to grasp synthesis, trajectory planning remains a critical component of robotic manipulation. Zhang et al. \cite{zhang2021trajectory} proposed a time-optimal trajectory planning strategy that incorporates dynamic constraints and input shaping algorithms to improve motion speed and smoothness. Jose et al. introduced a screw theory-based method for generating singularity-free waypoints, validated through experiments on parallel manipulators. 
Finally, Ortenzi et al. \cite{ortenzi2019singularity} developed an iterative method for determining joint solutions of redundant manipulators performing telemanipulation tasks. Their approach avoids singularities and joint limits, enabling smooth and reliable motion execution. These contributions collectively demonstrate the importance of integrating vision-based perception with robust motion planning schemes to enhance the adaptability, precision, and safety of robotic manipulators in complex environments.

Existing research in SDLs has made notable progress in robotic manipulation, motion planning, and vision-based perception, yet several key limitations persist. Many frameworks lack real-time adaptability in dynamic lab environments, particularly when interacting with textured or moving objects. Vision systems are often limited to static object detection or rely on depth sensors, without integrating feature-based tracking and homography-driven depth estimation for textured surfaces. Furthermore, while screw theory has been widely applied for kinematic modeling, its use in deriving both forward and inverse kinematics with stability guarantees under redundancy and singularities is still underexplored. The screw theory with the Damped Least Squares (DLS) method addresses this gap by enabling smooth and continuous joint trajectories. In motion planning, although algorithms like RRT* are known for optimality, their coupling with real-time visual feedback for adaptive trajectory generation remains limited in current literature. The proposed framework bridges this gap by combining RRT*-based planning with a vision pipeline that supports dynamic pose estimation and grasping. Additionally, few studies offer a unified simulation-based evaluation of these components using quantitative metrics such as RMSE pose errors, velocity continuity, and higher-order motion profiles. By addressing these gaps, our work contributes a cohesive and scalable solution for autonomous experimentation in next-generation SDLs.

\section{Methodologies}
\label{sec:Methodologies}

We performed kinematic analysis to model the motion constraints of the tracking system and assess the robot’s operational reach. This included solving inverse kinematics to determine joint configurations that enable the system to accurately follow an object’s path, thereby supporting effective grasping and manipulation.
We implemented a structured planar pose estimation algorithm for object tracking, which unfolds through several key stages. The process begins with extracting and matching features to reliably identify distinct characteristics of the target object. This is followed by homography estimation and perspective transformation, which establish spatial correspondence and alignment. Directional vectors are then computed on the object’s surface to analyze its orientation. Finally, depth data is integrated to estimate the object’s planar pose, ensuring precise interaction and grasping capabilities.
For object detection, the system first identifies salient features such as edges, corners, blobs, and ridges from images of planar objects using SIFT \cite{wu2013comparative}. These features are then matched with those captured by the camera to determine object presence. Once converted into feature vectors, matches are evaluated to confirm detection. To address the computational demands of high-dimensional feature matching, we utilized the FLANN-based K-d Nearest Neighbor Search \cite{muja2009fast}, which offers efficient performance for real-time applications.
Homography estimation relies on the matched features, though some may be incorrect and introduce noise. To overcome this, we applied the RANSAC algorithm \cite{fischler1981ransac}, which filters out false matches by iteratively selecting inliers from minimal subsets of data. This approach improves both speed and accuracy, making it ideal for dynamic environments. Perspective transformation is then used to estimate corresponding points in the test image, allowing us to derive basis vectors on the object’s surface. Depth information is subsequently incorporated to calculate the surface normal, enabling accurate 3D pose estimation of the planar object.

\subsection{Approach Overview}
\begin{figure}[hbt!]
    \centering
    \includegraphics[width=5 in, height = 1.5 in]{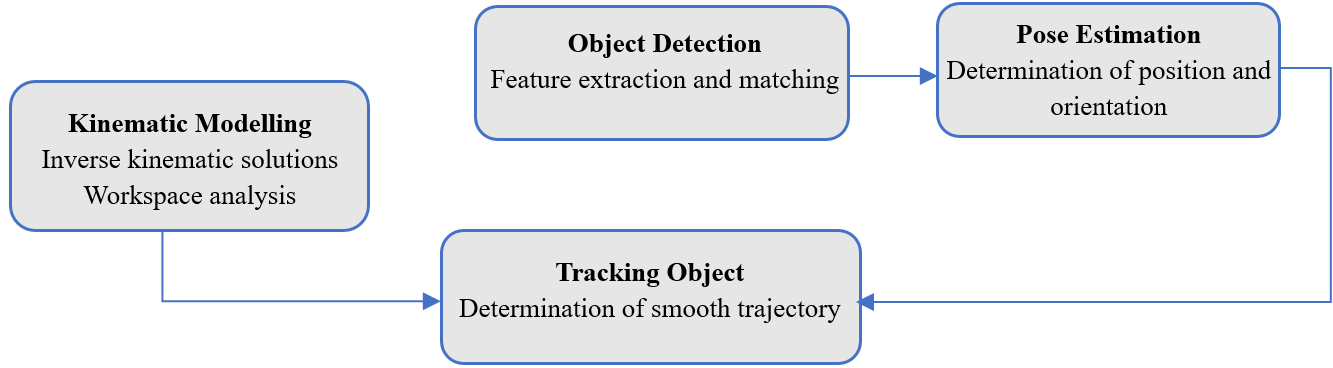}
    \caption{Methodology adopted in this work.}
    \label{flowchart}
\end{figure}

This work employs a KUKA LBR iiwa 14 manipulator and Robotiq Hand-E gripper to evaluate the developed vision based motion planning scheme as shown in Fig \ref{flowchart} . The kinematic equations governing the manipulator are derived using the screw theory approach. These equations facilitate the analysis of the manipulator's Cartesian workspace, which is essential for motion planning. Subsequently, motion planning is conducted to identify an optimal set of joint solutions for navigating trajectories within the manipulator and gripper workspaces. Trajectories are generated using the RRT* algorithm, while joint solutions are determined using the Damped Least Square (DLS) method. Although the simulation environments used in this study are obstacle-free, the RRT* algorithm was deliberately chosen over simpler interpolation methods to ensure generalization to more complex, cluttered environments typical of real-world SDL. RRT* offers asymptotic optimality and the ability to handle high-dimensional configuration spaces with dynamic constraints, making it well-suited for future extensions involving obstacle avoidance, constrained motion, and workspace reconfiguration. By integrating RRT* into the current framework, we establish a scalable and modular planning backbone that can be readily adapted to more realistic scenarios without requiring fundamental changes to the motion planning architecture.  

The kinematic equations obtained through the screw theory method enable the derivation of a set of feasible solutions.
A unique set of configurations is selected for traversing a given trajectory using the DLS method. To ensure smooth and optimal trajectories, an objective function based on the accelerations of Cartesian motions is incorporated alongside the trajectory function. The smoothness of trajectory motions is assessed using this function. 
Along with the smoothness function, some other characteristics of the joint motions such as velocities, accelerations, jerk and snap values are evaluated.
To further validate the performance of motion planning schemes, errors along the X, Y, and Z directions along with orientation errors of trajectory waypoints are compared. Simulation studies are conducted to evaluate the efficacy of each method in determining joint solutions for trajectories traversing various locations within the workspace. Building on this framework, object detection is achieved through feature extraction and matching, allowing the system to accurately identify and localize target objects within the environment. Once an object is detected, pose estimation determines its precise position and orientation, supplying essential spatial data for manipulation. In the final stage, adaptive tracking enables the robot to dynamically refine its motion strategy, mitigating singularities and ensuring stable interaction with the object. This sequential pipeline supports robust and flexible autonomous handling across complex and dynamic scenarios.



%
\subsection{Kinematic modeling of the manipulator}
\begin{figure}[hbt!]
    \centering
    \includegraphics[width=2.1 in, height = 3.5 in]{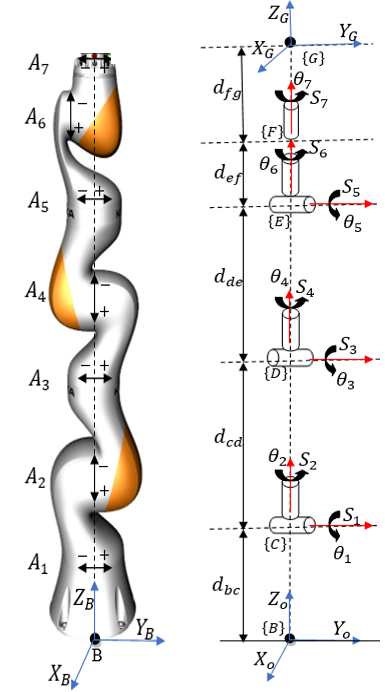}
    \caption{Illustration of various frames and screw axes of the used KUKA iiwa manipulator.}
    \label{fig:1}
\end{figure}

The kinematic modeling of the robot establishes a relationship between the angular variations of the manipulator's joints and the pose of the gripper. As stated before, the screw theory approach \cite{pardos2021screw} has been utilized for this purpose. The corresponding kinematic frames of the manipulator are shown in Fig. \ref{fig:1}. In this formulation, two inertial frames are employed to determine the poses of the end effector. One frame is designated to the manipulator base, and the other to the end effector joint. In this method, the displacement of rigid bodies is represented as the motion of a screw with a defined pitch. First, we present the modeling for the robot, followed by the modeling for the hand.
As shown in Fig. \ref{fig:1}, all the joints of the robot are assigned with a screw axis with zero pitch for the rotary joints. $A_{i}$  and $d_{ij}$ represent various manipulator joints and links, respectively. Link lengths of the used KUKA iiwa manipulator are given in Table \ref{tab:1}.
%
%
\begin{table}
\centering
  \caption{Manufacturer provided link dimensions for the KUKA iiwa 14 manipulator.}
  \label{tab:1}
    \begin{tabular}{c|c}
     \toprule
        Link & Dimension(mm)\\
        \midrule
        $d_{bc}$ & 340\\
        \midrule
        $d_{cd}$ & 740 \\
        \midrule
        $d_{de}$ & 400 \\
        \midrule
        $d_{eg}$ & 126 \\
        \midrule
        $d_{gf}$ &126 \\
    \bottomrule 
    \end{tabular}
\end{table}
\midsepdefault
Rodrigues formula for finding rotation matrix is given in \eqref{eq:1}
\begin{equation}
    \bm{R}(\hat{\omega},\theta) =e^{[\hat{\omega}]\theta} =\mathbf{I} + \mathrm{sin} \theta[\hat{\omega}] + (1 - \mathrm{cos}\theta)[\hat{\omega}]^2
    \label{eq:1}
\end{equation}
where $\hat{\omega}$ represents the $3\times 3$ skew-symmetric matrix of angular velocity matrix, $\omega$. $\theta$ denotes the angles of respective joints. Let $S_i$ represents the screw axes of various joints and 4x4 transformation matrix, $e^{[S]\theta}$ obtained using screw theory is given in \eqref{eq:2}. 

\begin{equation}
    e^{[S]\theta}=
    \begin{bmatrix}
    e^{[\hat{\omega}]\theta} & \mathbf{I}\theta + (1 - cos\theta)[\hat{\omega}] + (\theta - sin\theta)[\hat{\omega}]^{2}\mathbf{v} \\
    0 & 1
    \end{bmatrix}
     \label{eq:2}
\end{equation}
where, $\mathbf{v}$ is the 3x1 linear velocity vector and $\mathbf{I}$ is the 3x3 identity matrix. Final transformation matrix $\mathbf{T}$$(\theta)$ from base frame to end effector frame with $n$ number of joints is given in \eqref{eq:3}.

\begin{equation}
\mathbf{T}(\theta) =  e^{[S_1]\theta_1}e^{[S_2]\theta_2}......e^{[S_n]\theta_n}\mathbf{N}
\label{eq:3}
\end{equation}
where $\mathbf{N}$ is a 4 x 4 transformation matrix, which defines the initial pose of end effector frame with respect to base frame. Transformation matrix for the used 7-dof manipulator, $\mathbf{T}(\theta)_m$ is given in \eqref{eq:4}.
\begin{equation}
\mathbf{T}(\theta)_m =  e^{[S_1]\theta_1}e^{[S_2]\theta_2} e^{[S_3]\theta_3}e^{[S_4]\theta_4}e^{[S_5]\theta_5}e^{[S_6]\theta_6}e^{[S_7]\theta_7}  \mathbf{N}_m
\label{eq:4}
\end{equation}
where, 
\begin{equation}
    \mathbf{N}_m=
\begin{bmatrix}
1 & 0 & 0 & 0\\
0 & 1 & 0 & 0\\
0 &0 &1 &D\\
0 &0 &0 &1 
\end{bmatrix}
\label{eq:5}
\end{equation}
Distance between base frame and end effector frame, $D$ is given in \eqref{eq:6}.

\begin{equation}
     D=d_{bc}+d_{cd}+d_{de}+d_{ef}
\label{eq:6}   
\end{equation}
\subsection{Workspace analysis}
\begin{figure}[!h]
    \centering
    \includegraphics[width=5 in, height=3.8 in]{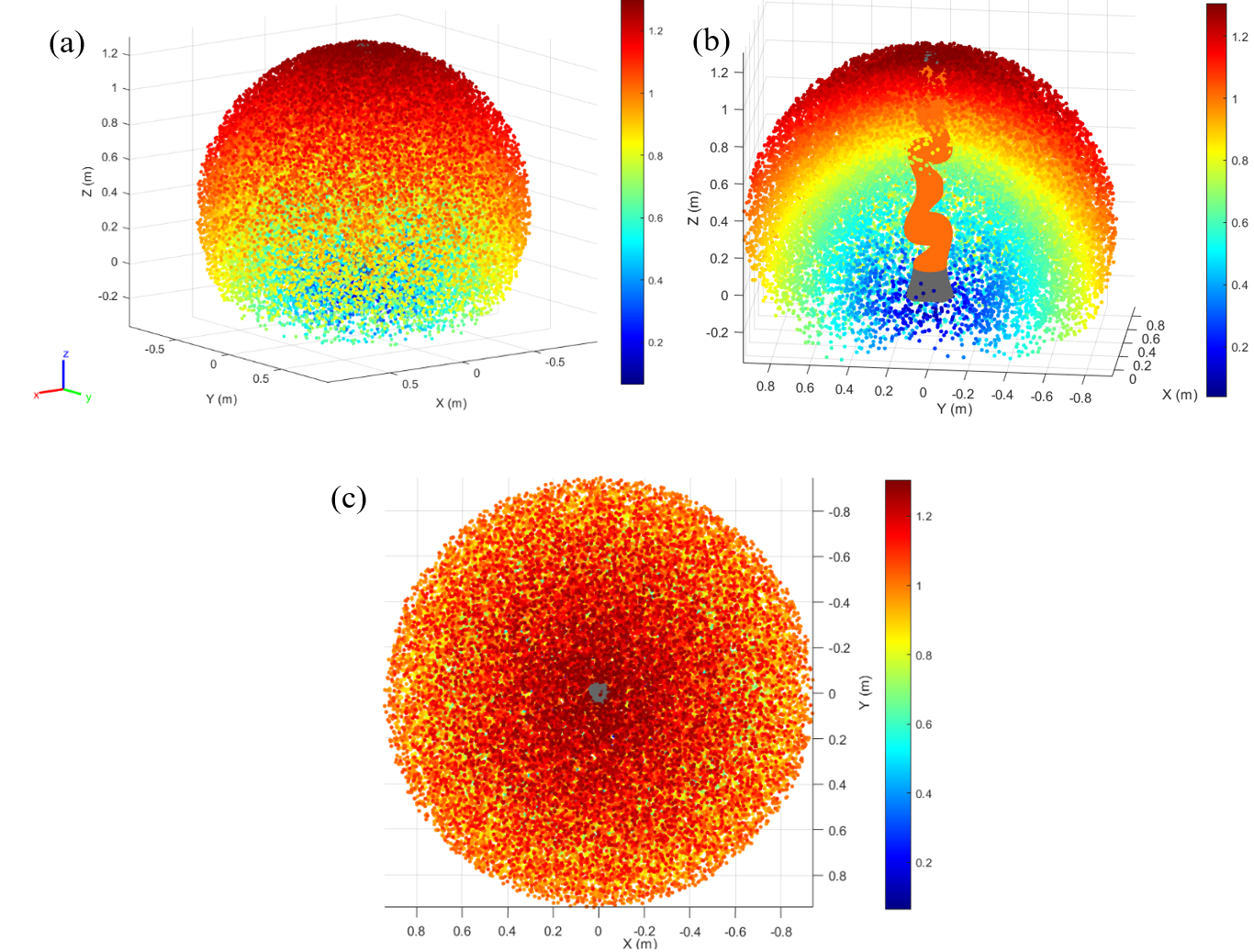}
    \caption{Views of combined workspace of the manipulator with hand (a)3D (b)Sectioned (c)Top. 
    }
    \label{fig:3}
\end{figure}

An analysis of the workspace was undertaken to identify singularities and unoccupied areas within the manipulator's operational space. Identifying distinct regions within the workspace facilitated the optimization of motion planning for various tasks. The combined Cartesian workspace of the manipulator and hand was determined using equation \eqref{eq:4}, considering joint limits. Fig.\ref{fig:3} illustrates views of the  workspace of the manipulator, showing all feasible positions of the manipulator and hand. This visualization accounted for joint limitations. Figs.\ref{fig:3} (a), (b), and (c) represent 3D, sectioned, and top views respectively of the cartesian workspace. Notably, the maximum reachable point within the workspace was found to be 0.78 meters. Moreover, the manipulator's dexterous workspace volume, including the gripper, was calculated to be 1.93 cubic meters. The obtained workspace serves as a basis for identifying the reachable areas of the manipulator during trajectory planning. By leveraging this information, any unreachable spaces traversed by the trajectory were excluded to prevent issues during the motion planning scheme.

%
\subsection{Determining Jacobian via screw theory}
The space Jacobian matrix, \(J_s(\theta)\), in fixed (space) frame coordinates, is a function of the joint angles \(\theta\). Given the joint velocities \(\dot{\theta}\), the task space (Cartesian) velocity \(V_s\) of an end-effector is given in \eqref{eq:13}
\begin{equation}
    V_s = J_s(\theta) \cdot \dot{\theta}
    \label{eq:13}
\end{equation}
where \(J_s(\theta)\) is the \(6 \times n\) Jacobian matrix, with \(3 \times n\) linear Jacobian elements, \(J_v\), and \(3 \times n\) angular Jacobian elements, \(J_\omega\) as given in \eqref{eq:14}.
\begin{equation}
    J_s (\theta) = 
    \begin{bmatrix}
        J_v\\
        J_w
    \end{bmatrix}
    \label{eq:14}
\end{equation}
Expanding \eqref{eq:13} for the 7-dof manipulator model, we obtain \eqref{eq:15} 
\begin{equation} 
    \begin{bmatrix}
    \dot x\\
    \dot y\\
    \dot z\\
    \omega_x\\
    \omega_y\\
    \omega_z
    \end{bmatrix}
    = \frac {dh(\theta)} {d(\theta)}
    \begin{bmatrix}
       \dot \theta_1\\
       \dot \theta_2\\
       \dot \theta_3\\
       \dot \theta_4\\
       \dot \theta_5\\
       \dot \theta_6\\
       \dot \theta_7 
    \end{bmatrix}
    \label{eq:15}
\end{equation}
where, $\frac {dh(\theta)} {d(\theta)}$ is given in equation \eqref{eq:16}
\begin{equation}
    \frac {dh(\theta)} {d(\theta)} =
    \begin{bmatrix}
        J_{11} &J_{12} &. &. &. &J_{17} \\
        J_{21} &J_{22} &. &. &. &J_{27} \\
        : &... &: &: &: &:\\
        . &... &: &: &: &:\\
       J_{61} &... &: &: &: &J_{67}
        \label{eq:16}
    \end{bmatrix}
\end{equation}
The space Jacobian, \(J_s(\theta)\), for the KUKA iiwa manipulator with \(n = 7\) joints, is determined using screw theory formulation given in \eqref{eq:17}.
\begin{equation}
   J_s (\theta) = [J_s (\theta)_1J_s (\theta)_2J_s (\theta)_3...........J_s (\theta)_n ]
   \label{eq:17}
\end{equation}
where each element \(J_s(\theta)_{i}\) represents the \(6 \times 1\) Jacobian of each joint. The Jacobian of the first joint, \(J_s(\theta)_1\), is equal to the screw axis, \(S_1\), as given in \eqref{eq:18}.
\begin{equation}
    J_s (\theta)_1 = S_1
    \label{eq:18}
\end{equation}
Jacobian of other joints are computed using \eqref{eq:19}.
\begin{equation}
    J_{s} (\theta)_i =[Ad_{{{e}^{[S_1]}......{e}^{[S_i-1]\theta_{i}-1]}}} ~ S_i ,
   ~ for ~ i = 2,3,...n
    \label{eq:19}
\end{equation}
%
%
%

\subsection{Inverse Kinematics Techniques - Damped Least Squares (DLS) Strategy}

Inverse kinematics (IK) algorithms are essential for determining joint parameters that guide a multi-link robotic system to a specified target pose. These methods compute joint angles that reposition the end-effector from its current location $P$ to a desired target $T$. The positional error is defined is given in \eqref{eq:24}:
\begin{equation}
    \Vec{e} = T - P = 0
    \label{eq:24}
\end{equation}
To minimize this error, the joint configuration $\theta$ is iteratively updated using \eqref{eq:27}:
\begin{equation}
    \Vec{e} = J_s(\theta)\Delta\theta
    \label{eq:27}
\end{equation}
The Jacobian matrix $J_s(\theta)$, derived via screw theory as outlined in \cite{lynch2017modern}, is central to all three IK strategies discussed below. However, solving \eqref{eq:27} may not always be feasible, especially when the manipulator possesses redundant degrees of freedom (DoFs). In singular configurations, the Jacobian becomes non-invertible, leading to an infinite number of solutions. To address this, alternative Jacobian formulations, as described in \cite{buss2004introduction}, can be employed. These alternatives improve performance by mitigating oscillations and overshoot but may occasionally produce non-feasible or suboptimal solutions in terms of energy efficiency and computational time. In singular configurations, the Jacobian is non-invertible, resulting in having infinite solutions. To avoid this issue, alternative Jacobians can be used as described in \cite{buss2004introduction}. These can improve performance by reducing oscillation and overshoot. However, they may sometimes result in non-feasible or non-optimal solutions in terms of energy and computational time. 

The DLS method seeks to minimize both the Cartesian error and the joint velocity norm. It is particularly effective near singularities and for underactuated systems. The objective function with a constant parameter, $W$ is given in \eqref{eq:41}:

\begin{equation}
    \min \| V_s - J_s\dot{\theta} \|^2_{W_1} + \| \Delta \theta \|
    \label{eq:41}
\end{equation}
The solution is obtained as given in \eqref{eq:42}:
\begin{equation}
    \dot{\theta} = J_s^{+}V_s
    \label{eq:42}
\end{equation}
where the damped inverse $J^+$ is defined as given in \eqref{eq:43}:
\begin{equation}
    J_s^{+} = (J_s^{T}W_{1}J_s + W_{2})^{-1}J_s^{T}W_{1}
    \label{eq:43}
\end{equation}
Here, $W_1 = I$ and $W_2 = \alpha I$, with $\alpha$ being a damping coefficient that balances stability and responsiveness. The weighting matrices $W_1$ and $W_2$ are positive definite matrices and non-singular in nature. The weighting matrix $W_1$ can be chosen to add priority to the task vector. This is very important in redundancy resolution approaches, as might alternatively be chosen to give the cost function energy units and so produce consistent results regardless of units or scale. 
The formulation given in \eqref{eq:43} works well for a rank-deficient Jacobian and can avoid generating high joint velocities near singularity regions. It blends joint velocity damping with minimizing the inaccuracy in following the given trajectory, resulting in a trade-off between the viability of the joint velocities and the precision with which the intended end-effector trajectory is followed. The joint update is computed using \eqref{eq:44}
\begin{equation}
    \Delta \theta = [J_s^{T}(J_sJ_s^{T} + \alpha^{2}I)^{-1}]_{7 \times 6} \Vec{e}_{6 \times 1}
    \label{eq:44}
\end{equation}
When $\alpha = 0$, the method simplifies to the pseudoinverse approach. Larger values of $\alpha$ reduce joint velocity magnitudes but may introduce trajectory deviations.  A larger value of $\alpha$ indicates a reduced joint velocity norm, although there can still be some deviation from the intended end-effector trajectory. Given that the DLS formulation must weigh the viability of the inverse kinematic solution against its precision, it is evident that the damping factor $\alpha$ is crucial. Therefore, it is important to set this parameter's value appropriately to guarantee that joint velocities are feasible in all configurations. 

\subsection{Vision-Based Grasping Framework}
To enable reliable object manipulation, we implemented a vision-guided grasping method \cite{paul2021object} that integrates perceptual input with adaptive planning strategies. The architecture comprises two core modules:

\begin{itemize}
  \item Planar object localization and pose inference
  \item Grasp synthesis based on dynamic pose updates
\end{itemize}

\subsubsection{Planar Object Localization and Pose Inference}

Robust pose estimation is a prerequisite for effective manipulation. Our method follows a four-stage pipeline:

\begin{enumerate}
  \item Extraction of visual features and descriptor matching
  \item Homography-based transformation and perspective alignment
  \item Derivation of object-centric coordinate axes
  \item Pose refinement using depth-enhanced feedback
\end{enumerate}

The pose estimation pipeline fuses 2D homography with RGB‑D measurements to recover the full 6‑DoF pose of planar, textured objects. First, the RGB image is processed with SIFT to extract keypoints and descriptors that are robust to viewpoint and scale changes. Descriptor matching against a pre-registered template is performed using FLANN, and RANSAC is subsequently applied to reject outliers and compute a homography that maps template coordinates to the current image plane. The homography is used to localize 2D corner (or keypoint) pixel coordinates in the RGB image, and then the depth image is generated at those pixel coordinates to obtain metric 3D points that are used for pose estimation. The homography is used purely for 2D localization: it projects the four template corners to pixel coordinates in the current frame, yielding high-confidence 2D points that anchor the object’s footprint. Depth fusion is then performed at these localized pixels. The RGB and depth frames are hardware-aligned using the camera intrinsics and distortion parameters obtained from calibration. For each corner pixel, the corresponding depth value is obtained from the synchronized depth image. These values are back-projected into 3D camera coordinates via the pinhole model, producing a set of 3D corner points that transform the 2D homography correspondences into metric space. If any corner depth is invalid (missing or beyond sensor range), nearest-neighbor interpolation over the local depth patch or planar fitting on inliers is used to recover a consistent 3D estimate; frames with insufficient valid corners are discarded to preserve reliability. The object pose is constructed from these 3D corners. The centroid defines the translation, while orientation is obtained by fitting a local orthonormal frame to the planar surface: the surface normal is computed from cross products of non-collinear corner edges, and in-plane axes are derived by orthogonalizing one edge direction against the normal. The resulting rotation matrix is refined by enforcing orthonormality. For numerical stability, a minimal set of geometric constraints is applied to reject degenerate configurations (nearly collinear corners or extremely acute aspect ratios), and the final pose is expressed either as Euler angles or a quaternion depending on downstream controller requirements. To mitigate measurement noise and ensure smooth control inputs, the estimated pose is temporally filtered before publication. 

All poses are transformed from the camera frame to the robot base frame using the fixed extrinsic calibration between the sensor and the manipulator, ensuring consistency with the kinematic chain. The final output is published as a  message comprising the filtered 6‑DoF pose and timestamp, which the motion stack consists of two phases: RRT* for the initial approach to the vicinity of the object, followed by DLS-based visual servoing that resolves incremental pose errors in real time without global re-planning. This design makes the roles of homography and depth explicit and complementary: homography provides robust 2D corner localization under texture and perspective changes, while depth supplies metric scale to recover 3D structure. Their fusion yields stable, accurate 6‑DoF poses suitable for dynamic manipulation, with clearly defined fallbacks, filtering, and frame transforms to maintain smooth end-effector trajectories.
Planar objects are detected using the SIFT algorithm, which identifies distinctive keypoints and encodes local image structure through descriptors. Matching is performed using FLANN for floating-point descriptors or Hamming distance for binary descriptors, establishing correspondences between the input image and a known template.
From the matched keypoints, a homography matrix $\mathbf{H}$ is computed to model the planar transformation as given in \eqref{v1}
\begin{equation}
\begin{bmatrix}
a \\ 
b \\ 
c
\end{bmatrix}
=
\mathbf{H}
\begin{bmatrix}
x \\ 
y \\ 
1
\end{bmatrix}
\label{v1}
\end{equation}
The projected coordinates $(x', y')$ are obtained using \eqref{v2}
\begin{equation}
x' = \frac{a}{c}, \quad y' = \frac{b}{c}
\label{v2}
\end{equation}
RANSAC is employed to eliminate outliers and ensure robust estimation.
To define the object’s local frame, three reference points are selected based on \eqref{v3}
\begin{equation}
P_c = (w/2, h/2), \quad P_x = (w, h/2), \quad P_y = (w/2, 0)
\label{v3}
\end{equation}
where $w$ and $h$ denote the object’s width and height. These points are projected into 3D using RGB-D data from a RealSense camera, yielding directional vectors using \eqref{v4} 
\begin{equation}
\vec{i} = \frac{\vec{x}}{\|\vec{x}\|}, \quad \vec{j} = \frac{\vec{y}}{\|\vec{y}\|}, \quad \vec{k} = \frac{\vec{x} \times \vec{y}}{\|\vec{x} \times \vec{y}\|}
\label{v4}
\end{equation}
The orthonormal basis $(\vec{i}, \vec{j}, \vec{k})$ defines the object’s orientation.
The rotation matrix $R$ is constructed from these vectors given in \eqref{v5}
\begin{equation}
R =
\begin{bmatrix}
i_x & j_x & k_x \\
i_y & j_y & k_y \\
i_z & j_z & k_z
\end{bmatrix}
\label{v5}
\end{equation}
Euler angles $(\phi, \theta, \psi)$ are derived as given in \eqref{v7}
\begin{equation}
\theta = \tan^{-1}(j_z), \quad \phi = \sin^{-1}(-i_z), \quad \psi = \tan^{-1}\left(\frac{i_y}{i_x}\right)
\label{v7}
\end{equation}
These angles encode the object’s spatial orientation and used for planning the grasping strategy.

\subsection{Motion evaluation metrics}

In this section, we describe the motion planning evaluation metrics adopted for analysing the motion smoothness. The evaluation metrics encompassed both motion smoothness and trajectory tracking performance. Motion smoothness was assessed using velocity, acceleration, jerk, and snap profiles, while tracking accuracy was quantified through Root Mean Square Error (RMSE) of the end-effector trajectory. To analyze dynamic consistency, velocity continuity and higher-order motion profiles were computed by evaluating the maximum differences between successive joint values across the trajectory.

After finalizing the trajectory waypoints, the joint angles of the manipulator and gripper were provided to achieve the desired end-effector motions. Smoothness of joint motions was evaluated based on the rate of change of acceleration over time. The smoothness function \(S_{func}\), derived from the third derivative of position, serves as an effective metric for smoothness. Since smoothness is inversely proportional to the rate of change of acceleration, the reciprocal of \(S_{func}\) was used as a measure of smoothness.  The performance of above-mentioned methods with respect to x, y and z coordinates within a time period, $t$ was evaluated using a reciprocal of smoothness function \cite{kato2023comparison} given in \eqref{eq:22}. 
\begin{equation}
    S_{func} = \frac{1}{\sqrt{\frac{1}{2}\int_{t_1}^{t_2}(\frac{d^3 x}{dt^3})^2+(\frac{d^3 y}{dt^3})^2+(\frac{d^3 z}{dt^3})^2 dt \times \frac{(t_2-t_1)^5}{l^2}}}
    \label{eq:22}
\end{equation}
where, length of trajectory, $l$ is given in \eqref{eq:23}
\begin{equation}
    l= \sum_{i=1}^{n-1}\sqrt{\Delta x_i^2 +\Delta y_i^2 + \Delta z_i^2 }
    \label{eq:23}
\end{equation}
As the rate of change of acceleration increases, smoothness decreases. Therefore, smoother motions correspond to lower rates of change in acceleration. Additionally, positional errors at the waypoints of the end-effector were analyzed to assess the accuracy of the inverse joint solutions.
To provide a comprehensive measure of overall error, this work introduces an RMSE metric that combines both positional and orientation errors. To effectively represent these combined errors as a single measure, positional errors (x, y, z) and orientation errors were calculated separately, normalized, and then integrated into a unified metric. 
For position, the RMSE can be computed using \eqref{eq:rmsepos}:
\begin{equation}
\label{eq:rmsepos}
    RMSE_{pos}=\sqrt{((1/N)*\sum_{i=1}^{N}((x_i-\hat{x_i} )^2+(y_i-\hat{y_i} )^2+(z_i-\hat{z_i})^2)}
\end{equation}
For orientation errors, since there are three angles (roll, pitch and yaw), the RMSE for each angle is computed separately as given in \eqref{m1} - \eqref{m3}:
\begin{equation}
    RMSE_{roll}=\sqrt{((1/N)*\sum_{i=1}^{N}((\theta_{roll,i}-\hat{\theta}_{roll,i} )^2)}
    \label{m1}
\end{equation}
\begin{equation}
    RMSE_{pitch}=\sqrt{((1/N)*\sum_{i=1}^{N}((\theta_{pitch,i}-\hat{\theta}_{pitch,i} )^2)}
    \label{m2}
\end{equation}
\begin{equation}
    RMSE_{yaw}=\sqrt{((1/N)*\sum_{i=1}^{N}((\theta_{yaw,i}-\hat{\theta}_{yaw,i} )^2)}
    \label{m3}
\end{equation}
To represent the overall orientation error as a single metric in radians, we combined the individual RMSE values for roll, pitch, and yaw into a single orientation RMSE using \eqref{m4} :
\begin{equation}
    RMSE_{orient}=\sqrt{((1/3)*(RMSE_{roll})^2 + (RMSE_{pitch})^2+(RMSE_{yaw})^2)}
    \label{m4}
\end{equation}
\section{Results and Discussion}
\label{sec:experiments}
%
%
%

\begin{figure}[hbt!]
   \centering
   \includegraphics[width=\linewidth]{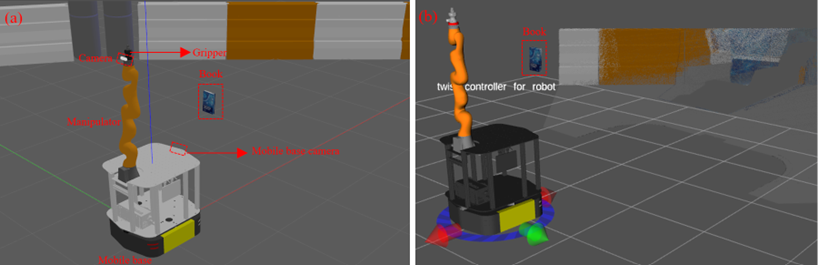}
    \caption{Simulation environment (a) Gazebo (b) Rviz. }
    \label{fig:5}
\end{figure}

The process of robotic manipulation in this work begins with kinematic modeling, which involved solving inverse kinematics and conducting workspace analysis to determine feasible configurations for the robot’s end effector. This foundational stage ensured that the manipulator can reach and interact with objects within its operational domain. Following this, object detection is performed through feature extraction and matching techniques, enabling the system to identify and localize target objects within the environment. Once detected, pose estimation is employed to determine the precise position and orientation of the object, providing critical spatial information for manipulation. The final stage involved tracking the textured object, in which the robot dynamically adjusts its position and orientation to follow the object. This sequential framework facilitates reliable and flexible autonomous handling in complex and variable settings.
The proposed object recognition and pose estimation algorithm were implemented within the Robot Operating System (ROS) framework, leveraging  OpenCV on an Ubuntu 20.04 platform. The system was deployed on a 3.0 GHz Intel Core i7-7400 CPU with 16GB of RAM. To evaluate the algorithm, a simulation environment was constructed using Rviz and Gazebo, featuring a mobile manipulator interacting with a textured book cover, as illustrated in Fig. \ref{fig:5}. The simulation setup, including both Gazebo and Rviz visualizations, is depicted in Figs. \ref{fig:5}(a) and \ref{fig:5}(b), respectively.
%
%

%

The motion strategies were evaluated based on the smoothness of joint motions using a smoothness function and the error range relative to the desired end-effector trajectory. Additionally, the velocity, acceleration, jerk, and snap values of the motions were determined to analyse the joint motions. The proposed method aimed to gradually adjust joint positions and orientations from a stable state. However, due to the arbitrary number of iterations, the time required to find an inverse kinematics (IK) solution for a given end-effector pose was variable. Despite this, the duration of a single iteration remained constant with respect to the dimensionality of \(J\) and \(\theta\), and was unaffected by the full algorithm's completion. To address this variability, a maximum time limit for the algorithm was enforced by setting an upper bound on the number of iterations. 
In Jacobian-based inverse kinematics solvers, the dimensionality of \(J\) in 3-dimensional space is typically either 3 or 6. A 3-dimensional \(J\) encodes only the positional information for the end-effector, while a 6-dimensional \(J\) is often preferred as it includes both positional and orientation information. In this work, a 6-dimensional \(J\) vector was chosen to account for both positional and orientation components. The system was deemed repeatable if a given goal vector \(\Vec{g}\) consistently produced the same pose vector. However, achieving repeatability in redundant systems requires special measures, as this consistency is not guaranteed inherently. An alternative approach involves resetting the system to a predefined default pose, ensuring repeatable solutions. However, this method may introduce sharp discontinuities in the solution trajectory. For every inverse solution technique employed in this work, the error matrix \(\Vec{e}\) was assigned values as described in \eqref{eq:47}:%
\begin{equation}
    \Vec{e} =
    \begin{bmatrix}
        X_{\mathrm{target}} - X_{\mathrm{endeffector}} \\
        Y_{\mathrm{target}} - Y_{\mathrm{endeffector}} \\
        Z_{\mathrm{target}} - Z_{\mathrm{endeffector}} \\
        \alpha_{\mathrm{target}} - \alpha_{\mathrm{endeffector}}\\
       \beta_{\mathrm{target}} - \beta_{\mathrm{endeffector}}\\
        \gamma_{\mathrm{target}} - \gamma_{\mathrm{endeffector}}
    \end{bmatrix} =
    \begin{bmatrix}
        0.01 \\
        0.01 \\
        0.01 \\
        0.01\\
        0.01\\
        0.01
    \end{bmatrix}
    \label{eq:47}
\end{equation}
$\Delta\Vec{e}$ was determined such that it moves $\Vec{e}$ closer to $\Vec{g}$. The starting iteration assumes $\Delta\Vec{e}$ as \(\Vec{g} - \Vec{e}\). The stopping conditions were implemented to improve the performance of the method and reduce computational effort during the iterations. In this work, the stopping criteria were as follows:
\begin{itemize}
    \item Finding a solution within the error limits given in \eqref{eq:48}.
    \item Convergence to a local minimum.
    \item Non-convergence after the allotted time.
    \item Maximum iterations reached.
\end{itemize}
If the solution converges to a local minimum, pose vectors can be randomized to avoid recurrence in future iterations. An allotted time can be specified for each step to prevent the method from exceeding a predetermined duration. Additionally, the maximum number of steps can be set to limit computational time. 
The iteration step size, \(\beta\), was determined using the desired \(\Vec{e}\) and \(\Vec{g}\) as given in \eqref{eq:48}:
\begin{equation}
    \Delta\Vec{e} = \beta (\Vec{g} - \Vec{e}), \quad 0 \leq \beta \leq 1
    \label{eq:48}
\end{equation}
The step size was limited by scaling it with \(\beta\). The determination of \(\beta\) was carried out after computing \(\Delta\theta\) for better approximation. A common approach involved limiting joint rotation increments to a maximum of 5 degrees per iteration. Initially, \(\Delta\theta\) was computed without including \(\beta\), and later checked to ensure no \(\Delta\theta\) values exceeded the threshold \(\beta_h\), calculated using equation \eqref{eq:49}:
\begin{equation}
    \beta_h = \frac{Threshold}{\max(Threshold, \max(|\Delta\theta_i|))}
    \label{eq:49}
\end{equation}
After finalizing \(\beta\), the \(\Delta\theta\) values were updated during iterations using \eqref{eq:50}:
\begin{equation}
    \theta = \theta + \beta (\Delta\theta)
    \label{eq:50}
\end{equation}

 The dampening constant of DLS method was determined by multiplying the size of the desired change in the end effector by a small constant that was set at each iteration. The small constant was taken as 1/100 of the total sum of the segments from base to end effector. Small constant was added to prevent the oscillating of errors due to desirable change was too small to dampen. The problem occurs because the target positions are nearly close to each other.  The inclusion of $\alpha$ avoids the convergence of jacobian matrix to a singular space at any time. An orthogonal set of vectors, which are equal to the length of the desirable change in vector, $\Vec{e}$ will result in a good damping effect. However, the effect resulted in increase in computational time for calculating the joint values. Each iteration tried to lower the value of errors to the desired $\Vec{e}$ value. The iteration stopped when the stopping criteria specified in section were met.

\begin{figure}[hbt!] \centering \includegraphics[width=0.5\textwidth]{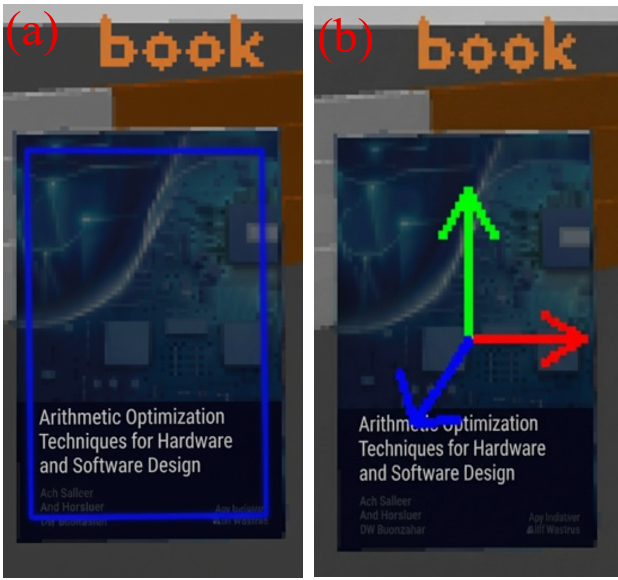} \caption{Vision algorithm outputs: (a) Bounding box, (b) Pose of object} \label{exp1} 
\label{b1}
\end{figure}
\begin{figure}[hbt!] \centering \includegraphics[width=1.0\textwidth, height=1.7 in]{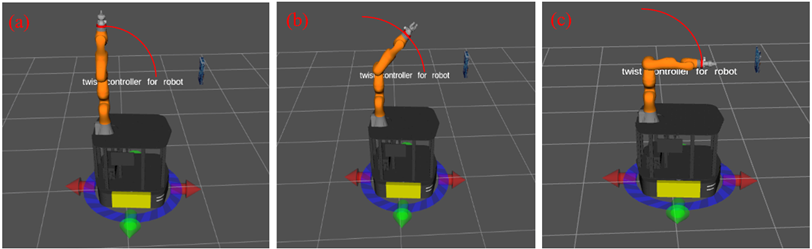} \caption{Manipulator motions to reach in front of a book in Rviz (a)Initial (b)Intermediate (c)Final} \label{b2} \end{figure}
\begin{figure}[hbt!] \centering \includegraphics[width=1.0\textwidth, height=2 in]{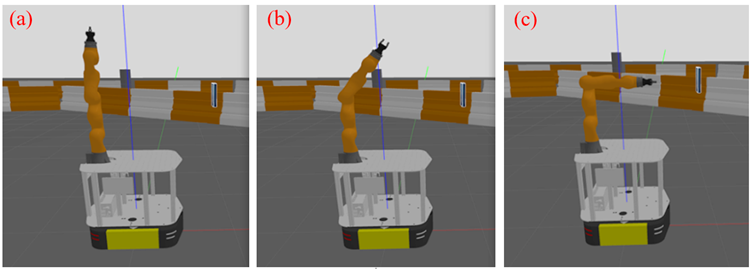} \caption{Manipulator motions to reach in front of a book in Gazebo (a)Initial (b)Intermediate (c)Final} \label{b3} \end{figure}

A RealSense camera mounted on the mobile base (shown in Fig. \ref{fig:5}) was employed to detect the initial pose of a textured book cover. Once the pose was identified, the RRT* motion planning algorithm utilized this information to generate a trajectory that allowed the manipulator to follow the book’s movement while maintaining a grasp-ready posture. DLS method was used to determine joint angles to traverse the trajectory.The outputs of the vision algorithm are presented in Figs.\ref{exp1}(a) and (b), where Fig.\ref{exp1}(a) shows the detected bounding box around the object, and Fig. \ref{exp1}(b) illustrates the estimated pose. These visual results confirmed the algorithm’s effectiveness in accurately localizing and orienting the target, providing essential input for subsequent manipulation.
The mobile manipulator was placed in front of a book as shown in Fig. \ref{fig:5}.  The manipulator follows a trajectory computed via the Rapidly-exploring Random Tree Star (RRT*) algorithm, with corresponding joint configurations derived using the Damped Least Squares (DLS) inverse kinematics method. The end-effector pose is defined such that the manipulator’s gripper approaches the book frontally, maintaining a face-to-face orientation at a distance of 15 cm in x-axis as shown in Figs. \ref{b2} and \ref{b3}. Figs. \ref{b2} and \ref{b3} illustrate the manipulator’s progression through its initial, intermediate, and final configurations as it approaches the target pose, visualized in Rviz and Gazebo environments, respectively.

\begin{figure}[hbt!] \centering \includegraphics[width=1.0\textwidth, height=1.7 in]{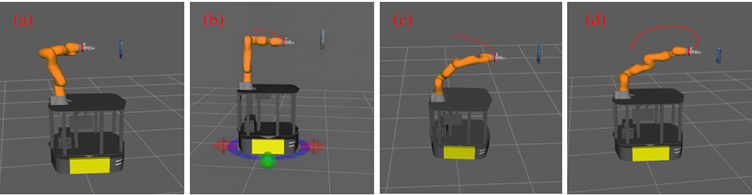} \caption{Manipulator motions to track a book in RviZ} \label{b4} \end{figure}

\begin{figure}[hbt!] \centering \includegraphics[width=1.0\textwidth, height=1.7 in]{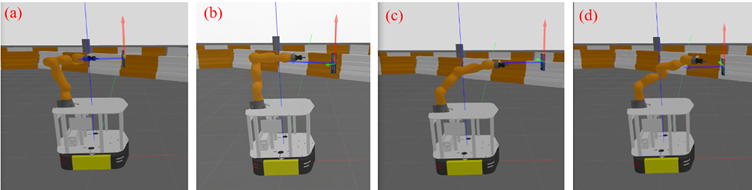} \caption{Manipulator motions to track a book in Gazebo} \label{b5} \end{figure}

To further validate the system’s manipulation capabilities, we performed tracking experiments following the motions of the book. During tracking, camera placed on the manipulator was used for detecting the updated poses of the book.
Screenshots of the manipulator’s motion in response to changes in the book’s pose in Rviz and Gazebo are shown in Figs. \ref{b4} – \ref{b5} respectively. These frames capture the adaptive motion planning and execution as the robotic arm reconfigures its joints and end effector to reach the target object. Mounted on a stationary mobile platform, the manipulator dynamically adjusts its configuration based on updated pose estimates from the vision system. The colored coordinate axes in each frame represent the estimated pose of the book, highlighting the system’s ability to track and align with the object throughout the approach. This visual evidence underscores the effectiveness of the integrated perception and planning framework in achieving precise and responsive object interaction.

The RRT* algorithm is employed exclusively during the initial approach phase to generate a collision-free, optimal trajectory from the manipulator’s current configuration to the vicinity of the target object. Once the end-effector reaches the location, the system transitions into a dynamic tracking mode, where continuous pose updates from the vision pipeline (at ~13 FPS) are used to compute incremental corrections via the DLS based inverse kinematics method. During this phase, the system is not re-planning full RRT* trajectories at each frame. Instead, the DLS controller operates in a visual-servoing loop, resolving the instantaneous pose error between the current end-effector position and the updated target pose. This separation between global planning (RRT*) and local tracking (DLS-based servoing) ensures computational efficiency while maintaining smooth and responsive motion.
These experiments demonstrated the framework’s ability to transition seamlessly from visual pose estimation and motion planning to physical interaction, validating its robustness in both tracking and manipulation tasks. All motion planning and execution were carried out within the ROS MoveIt framework, which ensured smooth and adaptive path optimization for precise object handling.
\begin{table}[htbp]
\centering
\caption{Performance Metrics for Object Tracking and Grasping}
\begin{tabular}{|l|c|p{8cm}|}
\hline
\textbf{Metric} & \textbf{Value} & \textbf{Description} \\
\hline
Tracking Accuracy & 96.7\% & Proportion of frames with accurate pose estimation prior to occlusion caused by end-effector constraints. \\
\hline
Pose Estimation Error & $\pm$0.63 cm & Mean spatial deviation between predicted and ground-truth object poses. \\
\hline
Detection Latency & 75 ms & Average per-frame processing time for object detection and pose update, supporting real-time operation. \\
\hline
Detection Precision & 97.1\% & Fraction of correctly identified objects among all detections. \\
\hline
Detection Recall & 96.5\% & Fraction of actual objects successfully detected. \\
\hline
Runtime Performance & $\sim$13.30 FPS & Average frame rate during continuous tracking and grasping. \\
\hline
\end{tabular}
\label{tab:performance_metrics}
\end{table}
\noindent
To obtain quantitative performance insights, we conducted a series of trials using a textured book cover placed at multiple positions within the camera’s field of view, under varying environmental conditions. The system achieved a tracking accuracy of 96.7\% as given in Table \ref{tab:performance_metrics}, indicating consistent pose estimation until occlusion occurred due to end-effector interference.  Pose estimation error remained within $\pm$0.63~cm, confirming the system’s suitability for precision grasping tasks. Pose estimation error (±0.63 cm) was computed by comparing the output of the vision pipeline to the ground-truth object pose provided by the Gazebo simulation environment. The detection pipeline maintained an average latency of 75~ms per frame, enabling real-time responsiveness at approximately 13.30~FPS. Detection precision and recall were measured at 97.1\% and 96.5\%, respectively, validating the system’s robustness in identifying and localizing target objects under challenging lighting and background conditions. These results affirm the system’s applicability to dynamic manipulation tasks in semi-structured environments, with strong generalization across object types and camera perspectives. The demonstrated performance highlights its potential for deployment in practical domains such as automated sorting, assistive robotics, and mobile manipulation.

\subsubsection{Evaluation of Trajectory Motions}


Maximum errors and RMSE of cartesian motions obtained using DLS method were shown in Table \ref{tab:5}.
\begin{table}[hbt!]
\midsepremove
 \caption{Maximum error and RMSE of Cartesian motions obtained using DLS method}
    \centering
    \begin{tabular}{c|c|c|c|c|c|c}
     \toprule
     \centering
             Error &x  &y &z &$\alpha$ &$\beta$ &$\gamma$ \\
           \midrule
        Maximum error(mm/deg)  &1.06	&1.11	&0.74  &1.75 &0.94 &1.05\\
        \midrule
         RMSE(mm/deg)  &0.97	&0.81	&0.66  &1.05 &0.71 &0.88\\

           \bottomrule 
     \end{tabular}
     \label{tab:5}
 \end{table}
 The maximum translational errors range from 0.74 mm to 1.11 mm, while rotational errors peak at 1.75 deg. The RMSE values indicate consistent performance, with translational deviations remaining below 1 mm and rotational deviations under 1.10 deg. These results demonstrate the DLS method’s effectiveness in maintaining low pose estimation errors across the trajectory.
Velocity Continuity (VC), Acceleration Profile (AP), jerk, snap, smoothness and RMSE errors of end effector were evaluated. Maximum values of AP,jerk, and snap values of motions were calculated to analyse the behaviour of motions. VC, AP, jerk, and snap values of manipulator joints are given in Table \ref{metrics} .
\begin{table}[hbt!]
\midsepremove
    \centering
    \caption{Velocity Continuity$(deg/s)$ , Acceleration Profile$(deg/s^{2})$ , jerk$(deg/s^{3})$ and snap$(deg/s^{4})$ values of manipulator joints}
    \centering
    \begin{tabular}{c|c|c|c|c|c|c|c}
     \toprule
     \centering
          Metrics  & $\theta_1$ &$\theta_2$ &$\theta_3$  & $\theta_4$ &$\theta_5$ &$\theta_6$  & $\theta_7$ \\
           \midrule
        VC &0.29	&0.24	&0.11	&0.30	&0.26	&0.08	&0.24	
           \\
          \midrule
       AP  &1.49	&1.08	&0.88
	&1.9	&1.58	&0.75	&1.88	\\
           \midrule
         Jerk  &0.39	&0.33	&0.28	&0.41	&0.30	&0.19	&0.37	\\
           \midrule 
         Snap   &0.46	&0.40	&0.37	&0.44	&0.32	&0.24	&0.43	\\
         \bottomrule
     \end{tabular}
    \label{metrics}
\end{table}
Table~\ref{metrics} presents the dynamic motion metrics of the manipulator joints, including velocity continuity (VC), acceleration profile (AP), jerk, and snap, evaluated across all seven joints ($\theta_1$ to $\theta_7$). The VC values range from 0.08~deg/s to 0.30~deg/s, indicating smooth transitions in joint velocities without abrupt discontinuities. The acceleration profiles span from 0.75~deg/s$^2$ to 1.9~deg/s$^2$, reflecting the rate of change in velocity during motion execution. Jerk values, which quantify the variation in acceleration, remain below 0.41~deg/s$^3$, suggesting well-regulated dynamic behavior. Similarly, snap values representing the fourth derivative of position are maintained below 0.46~deg/s$^4$, confirming the overall smoothness and stability of the joint trajectories. These metrics collectively demonstrate that the motion planning and control strategies employed yield dynamically consistent and mechanically safe joint movements suitable for precise and compliant manipulation tasks.

DLS performed better in terms of VC, AP, jerk and snap values of motions. Motions obtained using JT method exhibited lower VC, unstable AP, higher jerk and higher snap values. 
\begin{table}[hbt!]
\midsepremove
 \caption{Smoothness value of manipulator motions}
    \centering
    \begin{tabular}{c|c}
     \toprule
     \centering
          Joint  & Values   \\
          \midrule
           $\theta_1$ &0.78		
           \\
           \midrule
         
      $ \theta_{2}$ &1.57		\\
         \midrule
          $\theta_{3}$&0.81	 \\
           \midrule
          $\theta_{4}$  &1.03		 \\
           \midrule
         $\theta_{5}$  &0.86		\\
        \midrule
        $\theta_{6}$ &0.79		\\
        \midrule
        $\theta_{7}$ &2.14		\\
        \bottomrule
     \end{tabular}
     \label{tab:7}
 \end{table}
Table~\ref{tab:7} presents the smoothness values associated with the individual joint motions of the manipulator, denoted as $\theta_1$ through $\theta_7$.  The smoothness metric serves as a quantitative indicator of trajectory continuity, where lower values correspond to reduced dynamic fluctuations in joint motion specifically in velocity, acceleration, jerk, and snap thereby reflecting smoother and more mechanically stable trajectories. These values quantify the continuity and fluidity of joint trajectories during execution. The results indicate that most joints exhibit smooth motion profiles, with values ranging from 0.78 to 1.57. Notably, joint $\theta_7$ shows a higher smoothness value of 2.14, suggesting increased variability or dynamic complexity in its motion. Overall, the smoothness metrics reflect well-conditioned joint behavior, contributing to stable and precise end-effector control throughout the manipulation task.

 \subsection{Discussion}

 DLS method determined feasible joint motions with optimum accuracy and feasibility. It prevented the occurrences of high joint velocities and provide a smooth motion even near to singular regions by employing the appropriate damping factor.
 DLS method was employed to compute joint configurations for the manipulator, particularly in scenarios involving near-singular configurations and unreachable target poses. The core logic of the DLS approach involves assessing the contribution of each joint angle to the end-effector’s motion, comparing this influence to the remaining distance to the goal, and applying damping proportionally. When a joint’s influence significantly exceeds the required motion, its update is more aggressively damped. This is achieved by constraining the maximum allowable change in joint angles, thereby preventing abrupt or non-feasible motions.
In regions close to kinematic singularities, selecting an appropriate damping factor is critical. A poorly tuned damping constant can lead to excessive joint velocities or oscillatory behavior, especially when the manipulator attempts to reach unreachable targets. In such cases, the manipulator tends to extend toward the goal, approach a singular configuration, and oscillate due to the limitations of the Jacobian matrix, which provides only a first-order approximation of the end-effector’s motion. These higher-order effects become increasingly significant near singularities. To mitigate this, a linear model of the forward kinematics was constructed using screw theory. While effective in most regions, this model may lose accuracy near the workspace boundaries, where second-order corrections could be necessary.

Empirical tuning of the damping constant 
 revealed a trade-off between tracking precision and motion stability. Lower damping values minimized positional errors but introduced oscillations, while higher damping suppressed oscillations at the cost of accuracy. A compromise value was selected to balance these effects, resulting in smooth joint trajectories with tolerable error margins. The DLS method proved computationally efficient and capable of converging to feasible joint solutions without excessive effort.
Overall, the DLS approach demonstrated robustness in handling ill-conditioned Jacobians and singular configurations. The joint solutions obtained were considered optimal for the seven degrees of freedom of the manipulator, enabling stable and precise motion planning across diverse task scenarios.

 While the experiment utilized a textured book cover as a proxy object, the demonstrated capabilities particularly robust tracking of textured planar surfaces and smooth, redundant motion planning are directly transferable to high-value tasks in a SDL environment. 
 We acknowledge that the current vision pipeline, which relies on feature-based detection and homography-driven pose estimation, cannot generalize to more challenging SDL objects such as transparent glassware, reflective instrument panels, or low-texture plastic microplates. These object classes often lack sufficient visual features for reliable SIFT-based tracking and require alternative sensing modalities (e.g., depth fusion, polarization imaging, or thermal vision) or learning-based approaches (e.g., deep neural networks trained on domain-specific datasets) to ensure robust perception. However, the primary objective of this study is to establish a foundational motion and control framework that is modular and sensor-oriented. The architecture is designed to accommodate future upgrades to the perception stack without requiring changes to the underlying kinematic modeling or trajectory optimization methods. In this regard, the current implementation serves as a proof-of-concept that validates the motion planning and control pipeline under ideal visual conditions, providing a baseline for future extensions.



\section{Conclusion}
\label{sec:conclusion}

This study introduced a multimodal control framework that integrates vision-guided object tracking with Jacobian-based motion planning for autonomous manipulation in SDL environments. The proposed system combined feature-based detection and homography-driven pose estimation with RRT*-based trajectory generation and inverse kinematics computed via the Damped Least Squares (DLS) method. Kinematic modeling using screw theory enabled robust workspace analysis and stable joint solutions, even in the presence of redundancy and near-singular configurations.
Quantitative evaluations demonstrated the effectiveness of the framework. The manipulator achieved a maximum Cartesian pose error of 1.11~mm and a root mean square error (RMSE) of 0.97~mm across translational axes. Rotational RMSE remained below 1.10~deg, confirming precise orientation tracking. Joint motion profiles showed velocity continuity values ranging from 0.08~deg/s to 0.30~deg/s, with acceleration profiles peaking at 1.9~deg/s$^2$. Higher-order metrics such as jerk and snap were maintained below 0.41~deg/s$^3$ and 0.46~deg/s$^4$, respectively, indicating smooth and dynamically stable trajectories. Additionally, joint smoothness values ranged from 0.78 to 2.14, with most joints exhibiting consistent motion behavior.
Despite these promising results, certain limitations persist particularly in handling edge-of-workspace scenarios and highly nonlinear object motions. Future work will focus on enhancing the linear kinematic model with second-order approximations to improve accuracy near workspace boundaries. Real-time feedback from tactile and force-torque sensors will be incorporated to enable compliant manipulation and safe human-robot interaction. Furthermore, extending the framework to support multi-arm coordination and mobile base integration will enhance scalability and operational reach. Finally, benchmarking the system across diverse laboratory tasks and deploying it in real-world experimental settings will be essential for validating its generalizability and impact on autonomous scientific discovery. 

This work demonstrated foundational capabilities that align closely with the operational demands of a SDL. By showcasing reliable tracking of textured planar objects and smooth, redundant motion execution, the system lays groundwork for autonomous manipulation tasks central to SDL workflows. These include handling labeled reagent bottles, engaging with instrument interfaces, and managing microplate logistics. The use of a textured book cover, while simplified, effectively illustrated the system’s potential to generalize across SDL-relevant tasks. Future work will focus on enhancing the perception module to handle non-ideal laboratory objects by integrating multi-modal sensing and deep learning-based object detection and pose estimation. Additionally, we aim to evaluate the system’s performance in real-world SDL scenarios involving diverse object geometries, material properties, and environmental conditions. These extensions will further validate the scalability and robustness of the proposed framework in practical autonomous experimentation workflows.

%

\authorcontributions{
Conceptualization, methodology, software implementation, and manuscript preparation were carried out by Shifa Sulaiman. Amarnath A H contributed to simulations of work. Simon Bøgh and Naresh Marturi provided supervision, critical review, and guidance throughout the research and manuscript refinement process. All authors have read and agreed to the published version of the manuscript.
}

\funding{This work was supported by the Pioneer Center for Accelerating P2X
Materials Discovery, CAPeX.}


\conflictsofinterest{The authors declare no conflicts of interest.} 

\begin{adjustwidth}{-\extralength}{0cm}

\reftitle{References}

\bibliography{references}
\end{adjustwidth}

\end{document}